%% file: main.tex
\documentclass[10pt,twocolumn,letterpaper]{article}

\usepackage{cvpr}              %
\usepackage[accsupp]{axessibility}
\usepackage{graphicx}
\usepackage{amsmath}
\usepackage{amssymb}
\usepackage{mathtools}
\usepackage{booktabs}
\usepackage{amsfonts}
\usepackage{balance}

\usepackage[pagebackref,breaklinks,colorlinks]{hyperref}

\usepackage[capitalize]{cleveref}
\crefname{section}{Sec.}{Secs.}
\Crefname{section}{Section}{Sections}
\Crefname{table}{Table}{Tables}
\crefname{table}{Tab.}{Tabs.}

\def\cvprPaperID{8311} %
\def\confName{CVPR}
\def\confYear{2023}

\input{preamble}

\begin{document}

\title{Decoupling Human and Camera Motion from Videos in the Wild}

\author{Vickie Ye \qquad Georgios Pavlakos\qquad Jitendra Malik \qquad Angjoo Kanazawa \\
University of California, Berkeley
}

\twocolumn[
{%
\renewcommand\twocolumn[1][]{#1}%
\maketitle
    \begin{center}
    \vspace{-0.2in}
    \includegraphics[width=0.95\textwidth]{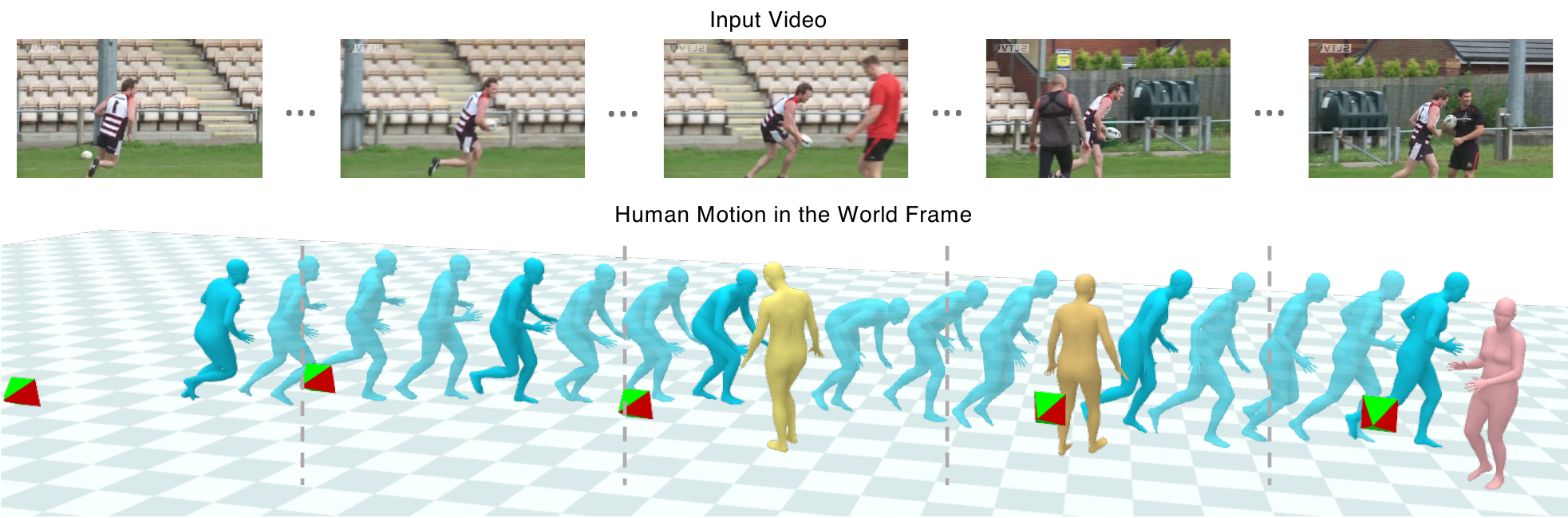}
    \vspace{-.5em}
    \captionof{figure}{
      \textbf{4D Reconstruction of People from Videos in-the-Wild.} We present \textbf{SLAHMR}: \textbf{S}imultaneous \textbf{L}ocalization \textbf{A}nd \textbf{H}uman \textbf{M}esh \textbf{R}ecovery, a method that given a video of moving people (top),
recovers the global trajectories of all people and cameras in the world coordinate space (bottom).
      We combine geometric insights, which determine relative camera motion,
      with learned human motion priors,
      which constrain a person's plausible displacement between frames,
      to position the people and cameras in the shared world frame through time.
      Our method can recover the global trajectories of all detected people from in-the-wild videos with uncontrolled camera and human motion.
      Please see the \href{https://vye16.github.io/slahmr}{project page} to see the full video results. 
    }
    \label{fig:teaser}
    \end{center}%
}]

\input{sec/0_abstract}
\section{Introduction}
\input{sec/1_intro}
\section{Related Work}
\input{sec/2_background}

\section{Method}
\input{sec/3_method}

\section{Experimental Results}
\input{sec/4_experiments}

\section{Discussion}
\input{sec/6_conclusion}

\balance
\newpage
{\small
\bibliographystyle{ieee_fullname}
\bibliography{egbib}
}

\appendix
\section*{\Large Appendix}
\input{sec/7_supp}

\end{document}

%% file: preamble.tex
\usepackage{overpic}
\usepackage{enumitem} %
\usepackage{overpic} %
\usepackage{color}
\usepackage{microtype} %
\usepackage{xcolor}

\definecolor{turquoise}{cmyk}{0.65,0,0.1,0.3}
\definecolor{purple}{rgb}{0.65,0,0.65}
\definecolor{dark_green}{rgb}{0, 0.5, 0}
\definecolor{orange}{rgb}{0.8, 0.6, 0.2}
\definecolor{red}{rgb}{0.8, 0.2, 0.2}
\definecolor{darkred}{rgb}{0.6, 0.1, 0.05}
\definecolor{blueish}{rgb}{0.0, 0.3, .6}
\definecolor{light_gray}{rgb}{0.7, 0.7, .7}
\definecolor{pink}{rgb}{1, 0, 1}
\definecolor{greyblue}{rgb}{0.25, 0.25, 1}

\newcommand{\bigP}{\mathcal{\mathbf{P}}}
\newcommand{\ti}[1]{#1^i_t}
\newcommand{\tf}[1]{\mathbf{#1}}
\newcommand{\w}{{}^\textrm{w}}
\newcommand{\cam}{{}^\textrm{c}}

%% file: sec/0_abstract.tex
\begin{abstract}
We propose a method to reconstruct global human trajectories %
from videos in the wild.
Our optimization method decouples the camera and human motion,
which allows us to place people in the same world coordinate frame. %
Most existing methods do not model the camera motion; methods that rely on the background pixels to infer 3D human motion usually require a full scene reconstruction, which is often not possible for in-the-wild videos.
However, even when existing SLAM systems cannot recover accurate scene reconstructions,
the background pixel motion still provides enough signal to constrain the camera motion.
We show that relative camera estimates along with data-driven human motion priors can resolve the scene scale ambiguity and recover global human trajectories.
Our method robustly recovers the global 3D trajectories of people in challenging in-the-wild videos, such as PoseTrack.
We quantify our improvement over existing methods on 3D human dataset Egobody.
We further demonstrate that our recovered camera scale allows us to reason about motion of multiple people in a shared coordinate frame,
which improves performance of downstream tracking in PoseTrack.

\end{abstract}

%% file: sec/1_intro.tex
Consider the video sequence in Figure~\ref{fig:teaser}. As human observers, we can clearly perceive that the camera is following the athlete as he runs across the field.
However, when this dynamic 3D scene is projected onto 2D images, because the camera tracks the athlete, the athlete appears to be at the center of the camera frame throughout the sequence
--- \ie the projection only captures the {\it net} motion of the underlying human and camera trajectory.
Thus, if we rely only on the person's 2D motion, as many human video reconstruction methods do,
we cannot recover their original trajectory in the world (Figure~\ref{fig:teaser} bottom left).
To recover the person's 3D motion in the world (Figure~\ref{fig:teaser} bottom right), we must also reason about how much the camera is moving.

We present an approach that models the camera motion
to recover the 3D human motion in the world from videos in the wild. %
 Our system can handle multiple people and reconstructs their motion in the same world coordinate frame, enabling us to capture their spatial relationships. 
Recovering the underlying human motion and their spatial relationships is a key step towards understanding humans from in-the-wild videos.
Tasks, such as autonomous planning in environments with humans~\cite{rudenko2020human},
or recognition of human interactions with the environment \cite{zhang2020perceiving} and other people \cite{knapp2013nonverbal, ng2020you2me},
rely on information about global human trajectories.
Current methods that recover global trajectories either require additional sensors, e.g. multiple cameras or depth sensors~\cite{guzov2021human,saini2022smartmocap},
or dense 3D reconstruction of the environment~\cite{liu20204d,PROX:2019},
both of which are only realistic in active or controlled capture settings.
Our method acquires these global trajectories from videos in the wild,
with no constraints on the capture setup, camera motion, or prior knowledge of the environment.
Being able to do this
from dynamic cameras is particularly relevant with the 
emergence of large-scale egocentric video datasets~\cite{Damen2022RESCALING, grauman2022ego4d, zhang2021egobody}.

To do this, given an input RGB video, we first estimate the relative camera motion between frames from the static scene's pixel motion with a SLAM system~\cite{teed2021droid}.
At the same time, we estimate the identities and body poses of all detected people with a 3D human tracking system~\cite{rajasegaran2022tracking}.
We use these estimates to initialize the trajectories of the humans and cameras in the shared world frame.
We then optimize these global trajectories over multiple stages
to be consistent with both the 2D observations in the video and learned priors about how human move in the world~\cite{rempe2021humor}.
We illustrate our pipeline in Figure~\ref{fig:pipeline}.
Unlike existing works~\cite{guzov2021human,liu20204d}, we optimize over human and camera trajectories in the world frame
without requiring an accurate 3D reconstruction of the static scene.
Because of this, our method operates on videos captured in the wild, a challenging domain for prior methods that require good 3D geometry, since these videos 
rarely contain camera viewpoints with sufficient baselines for reliable scene reconstruction.

We combine two main insights to enable this optimization.
First, even when the scene parallax is insufficient for accurate scene reconstruction,
it still allows reasonable estimates of camera motion up to an arbitrary scale factor.
In fact, in Figure~\ref{fig:pipeline}, the recovered scene structure for the input video is a degenerate flat plane,
but the relative camera motion still explains the scene parallax between frames.
Second, human bodies can move realistically in the world in a small range of ways.
Learned priors capture this space of realistic human motion well.
We use these insights to parameterize the camera trajectory
to be both consistent with the scene parallax
and the 2D reprojection of realistic human trajectories in the world.
Specifically, we optimize over the scale of camera displacement, using the relative camera estimates, to be consistent with the human displacement.
Moreover, when multiple people are present in a video, as is often the case in in-the-wild videos,
the motions of all the people further constrains the camera scale,
allowing our method to operate on complex videos of people.

We evaluate our approach on EgoBody~\cite{zhang2021egobody}, a new dataset of videos captured with a dynamic (ego-centric) camera with ground truth 3D global human motion trajectory. 
Our approach achieves significant improvement upon the state-of-the-art method that also tries to recover the human motion without considering the signal provided by the background pixels~\cite{yuan2022glamr}. %
We further evaluate our approach on PoseTrack~\cite{andriluka2018posetrack}, a challenging in-the-wild video dataset originally designed for tracking. To demonstrate the robustness of our approach, we provide the results on all PoseTrack validation sequences on our project page.
On top of qualitative evaluation, since there are no 3D ground-truth labels in PoseTrack, we test our approach through an evaluation on the downstream application of tracking. We show that the recovered scaled camera motion trajectory can be directly used in the PHALP system~\cite{rajasegaran2022tracking} to improve tracking. The scaled camera enables more persistent 3D human registration in the 3D world, which reduces the re-identification mistakes. %
We provide video results and code at the \href{https://vye16.github.io/slahmr}{project page}.

%% file: sec/2_background.tex
\begin{figure*}[t]
\centering
\includegraphics[width=\textwidth]{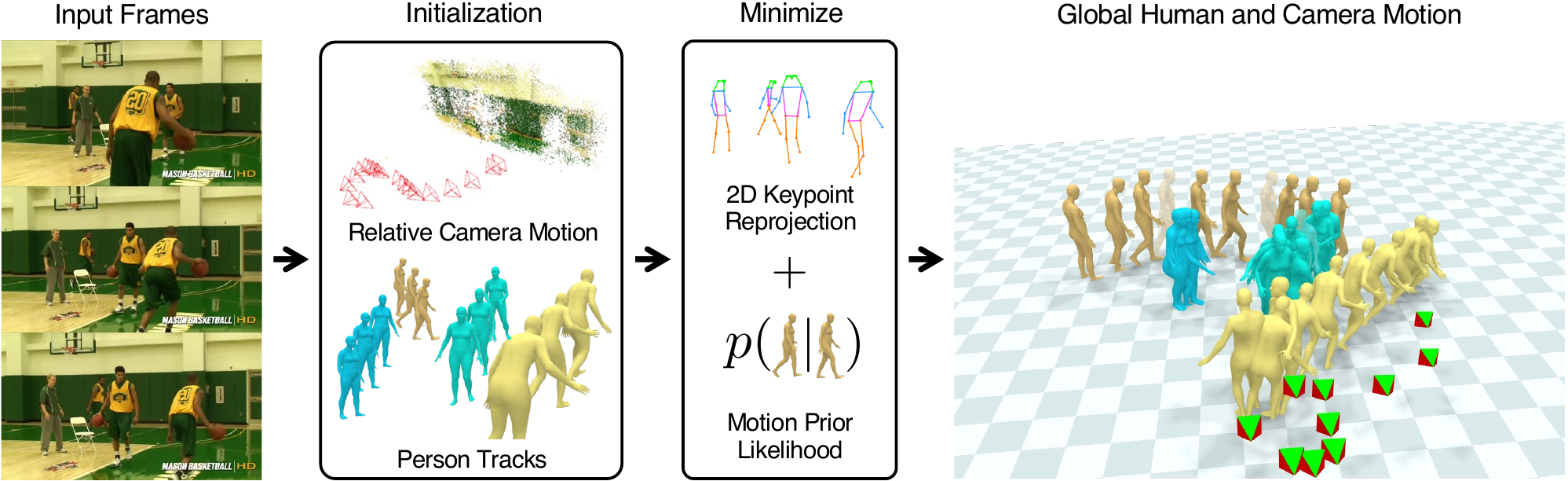}
\caption{
\textbf{SLAHMR Pipeline.} \looseness=-1 Given an input video in-the-Wild with moving camera (left), we first predict the relative camera motion with SfM~\cite{teed2021droid} (middle top). We also recover the unique identities of people across the video and their local 3D poses~\cite{rajasegaran2022tracking}. These are input into the proposed joint optimization system, which solves for the 4D trajectories of the moving people in the world coordinate frame, as well as the scale and the ground of the world. 
}
\label{fig:pipeline}
\end{figure*}

\paragraph{Human Mesh Recovery from a Single Image.}
In the literature for 3D human mesh reconstruction, most methods operate by recovering the parameters of a parametric human body model, notably  SMPL~\cite{loper2015smpl} or its follow-up models~\cite{osman2020star,pavlakos2019expressive, romero2017embodied,xu2020ghum}. 
The main paradigms are optimization-based, \eg, SMPLify~\cite{bogo2016keep} and follow-ups~\cite{lassner2017unite,pavlakos2019expressive,tiwari2022pose}, or regression-based, like HMR~\cite{kanazawa2018end} and follow-ups~\cite{guler2019holopose,kolotouros2019learning,zhang2021pymaf}.
For regression approaches in particular, many efforts have focused on increasing the model robustness in a variety of settings~\cite{Rockwell2020,joo2021exemplar,kocabas2021pare,kolotouros2021probabilistic,pavlakos2022human}. Most of these approaches predict the human mesh in the camera coordinate frame with identity camera.
There are recent works, \eg, SPEC~\cite{kocabas2021spec} and CLIFF~\cite{li2022cliff}, that also consider incorporating camera information in the regression pipeline, but only for single frame inference. 
PHALP~\cite{rajasegaran2022tracking} is a state-of-the-art method on tracking using the predicted 3D information of people ran on each frame. We use the detected identities and predicted 3D mesh as the initialization and show how it can be improved by incorporating the camera obtained by our approach.  %

\paragraph{Human Mesh Recovery from Video.}
Many works extend human mesh recovery approaches on video to recover a smooth plausible human motion.
However, these works fail to account for camera motion and do not recover global human trajectories.
Regression approaches like HMMR~\cite{kanazawa2019learning}, VIBE~\cite{kocabas2020vibe}, and follow-ups~\cite{choi2021beyond,pavlakos2022human,luo20203d} operate on a bounding box level and only consider the local motion of the person within that bounding box.
These approaches are prone to jitter since they are sensitive to the bounding box size. 
More recently, approaches such as GLAMR~\cite{yuan2022glamr}, D\&D~\cite{li2022d} and Yu~\etal~\cite{yu2021human},
have tried to circumvent the issue of camera motion by recovering plausible global trajectories from the per-frame local human poses.
However, relying {\it only} on local pose is not sufficient for a faithful global trajectory, especially for out-of-distribution poses,
and is brittle when local pose cannot be fully observed.
As such, \cite{yuan2022glamr} struggles on in-the-wild videos, which often have partial occlusions and diverse human actions.
Our work explicitly accounts for the camera motion
to place the humans in the static scene.

Optimization approaches are similarly limited by the lack of camera awareness.
\cite{arnab2019exploiting,pavlakos2022human} use body pose smoothness priors to recover net human motion over short sequences, ignoring cameras entirely.
Recent methods achieve more realistic human motion by modeling human dynamics,
through learned priors~\cite{rempe2021humor} or physics based priors~\cite{peng2018sfv, rempe2020contact,shimada2020physcap,PhysAwareTOG2021,yuan2021simpoe,xie2021physics,gartner2022trajectory}. 
These priors are naturally defined in the human coordinate frame, and have thus far been limited to settings where the camera is metrically known, or static. %
Our approach opens a path in which these methods can be applied to moving cameras. 

Other works rely on prior 3D scene information or additional sensors to contextualize human motion.
\cite{guzov2021human,pavlakos2022one,liu20204d} can recover faithful global trajectories when the cameras and dense 3D environment have already been reconstructed.
Such reconstructions require observations of the scene from many viewpoints with wide baselines.
\cite{guzov2021human, liu20204d} both rely on reconstructions from actively controlled capture data; \cite{pavlakos2022one} rely on television data in which the same set was observed from many different viewpoints.
\cite{saini2022smartmocap} recovers global human trajectories with multiple synchronized cameras, again only realistic for controlled capture settings, or a single static camera.
In contrast, our work recovers human trajectories for in-the-wild videos, in which camera motion is uncontrolled, and the scene reconstruction is limited or non-existent.
\cite{henning2022bodyslam} operate on monocular sequences, but the extent of results is limited to a single unoccluded person slowly walking in an indoor studio.
We demonstrate our approach on PoseTrack, a complex in-the-wild dataset, which includes videos with a large number of people in various environments. 

\paragraph{Human Mesh Recovery for Multiple People.}
There have been many works that consider the reconstruction of multiple people from single images. Zanfir~\etal~\cite{zanfir2018deep} propose an optimization approach, while follow-up work~\cite{zanfir2018deep,fieraru2021remips,zhang2021bmp} has considered regression solutions.
Jiang~\etal~\cite{jiang2020coherent} incorporate constraints that encourage the consistency of the multiple people in 3D using a Mask R-CNN~\cite{he2017mask} type of network, while Sun~\etal~\cite{sun2021monocular,sun2022putting} has investigated center-based regression~\cite{zhou2019objects}.
Mustafa~\etal~\cite{mustafa2021multi} consider implicit representations for the multiple person recovery.
However, all of the above works operate on a single frame basis.
Mehta~\etal~\cite{mehta2020xnect} operate on video but they only reconstruct the 3D skeleton and demonstrate results on simpler sequences with a static camera.
In contrast we recover the 3D trajectories of multiple people from a moving camera.

%% file: sec/3_method.tex
We take as input a video with $T$ frames of a scene with $N$ people.
Our goal is to recover the motion of all detected people in the world coordinate system.
We use the SMPL-H model~\cite{loper2015smpl,romero2017embodied} and represent each person $i$ at timestep $t$ via
global orientation $\ti{\Phi} \in \mathbb{R}^3$,
body pose (22 joint angles), $\ti{\Theta} \in \mathbb{R}^{22 \times 3}$,
shape $\beta^i \in \mathbb{R}^{16}$, shared over all timesteps $t$, 
and root translation $\ti{\Gamma} \in \mathbb{R}^3$:
\begin{equation}
    \ti{\bigP} = \{\ti{\Phi}, \ti{\Theta}, \beta^i, \ti{\Gamma} \}.
\end{equation}
The SMPL model uses these parameters to generate
the mesh vertices $\ti{\tf{V}} \in \mathbb{R}^{3 \times 6890}$
and joints $\mathbf{J}^i_t \in \mathbb{R}^{3 \times 22}$ of a human body
through the differentiable function $\mathcal{M}$:
\begin{equation} \label{eqn:obs_frame}
    [\mathbf{V}^i_t, \mathbf{J}^i_t] = \mathcal{M}(\Phi^i_t, \Theta^i_t, \beta^i) + \Gamma^i_t.
\end{equation}

We begin by estimating each person's per-frame pose $\ti{\hat{\bigP}}$
and computing their unique identity track associations over all frames
using state-of-the-art 3D tracking system, PHALP~\cite{rajasegaran2022tracking}.
PHALP estimates poses independently per-frame, and each estimate resides in the camera coordinate frame.
In a video, however, a person's motion in the camera coordinates is a composition of the human and camera motion in the world frame, \ie, the net motion:
\begin{equation}
    \textrm{camera motion} \circ \textrm{human motion} = \textrm{net motion}.
\end{equation}
To recover the original world trajectory of each person, we must determine the camera motion contribution to their net perceived motion.
We denote the pose in the camera frame as 
$\cam \ti{\bigP} = \{\cam \ti{\Phi}, \ti{\Theta}, \beta^i, \cam \ti{\Gamma} \}$,
and the pose estimate in the world frame as 
$\w \ti{\bigP} = \{\w \ti{\Phi}, \ti{\Theta}, \beta^i, \w \ti{\Gamma} \}$;
the local pose $\ti{\hat{\Theta}}$ and shape $\hat{\beta}^i$ parameters are the same in both.

Our first insight is to use the information in the static scene's pixel motion to compute the relative camera motion between video frames.
We use state-of-the-art data-driven SLAM system, DROID-SLAM~\cite{teed2021droid}
to estimate the world-to-camera transform at each time $t$, $\{\hat{R}_t, \hat{T}_t\}$. %
The camera motion can only be estimated up to an unknown scale of the world,
but human bodies and motion can only take on a plausible range of values in the world.
In order to ultimately place the people in the world, 
we must therefore determine $\alpha$,
the relative scale between the displacement of the camera and that of people.

Our second insight is to use priors about human motion in the world to jointly determine the camera scale $\alpha$
and people's global trajectories.
In the following sections, we describe the steps we take to initialize and prepare for joint optimization with
a data-driven human motion prior.
In Section~\ref{sec:stage1}, we describe how we initialize the multiple people tracks and cameras in the world coordinate frame.
In Section~\ref{sec:stage2}, we describe a smoothing step on the trajectories in the world, to warm-start our joint optimization problem.
Finally in Section~\ref{sec:stage3}, we describe the full optimization of trajectories and camera scale using the human motion prior.
\subsection{Initializing people in the world.}
\label{sec:stage1}
We take as input to our joint optimization problem
the pose parameters predicted by PHALP in the camera coordinate frame, $\ti{\hat{\bigP}}$,
and the world-to-camera transforms estimated with SLAM, $\{\hat{R}_t, \hat{T}_t\}$.
We initialize optimization variables $^\textrm{w}\ti{\bigP}$,
for all people $i=0,\dots,N-1$ and timesteps $t=0,\dots,T-1$.
The shape $\beta_i$ and pose $\ti{\Theta}$ parameters are defined in the human canonical frame, so we use PHALP estimates directly.
We initialize the global orientation and root translation in the world coordinate frame
using the estimated camera transforms and camera-frame pose parameters.
\begin{align}
    \w \ti{\Phi} &=
        R_t^{-1} \cam \ti{\hat{\Phi}}, &
    \w \ti{\Gamma} &=
        R_t^{-1} \cam \ti{\hat{\Gamma}} - \alpha R_t^{-1} T_t, \nonumber \\
    \beta_i &= \hat{\beta}^i, &
    \ti{\Theta} &= \ti{\hat{\Theta}},
\end{align}
where we initialize the camera scale $\alpha = 1$.
The joints in the world frame are then expressed as:
\begin{eqnarray} \label{eqn:world_frame}
\w\ti{\tf{J}} &=& \mathcal{M}(\w\ti{\Phi}, \ti{\Theta}, \beta^i) + \w\ti{\Gamma}.
\end{eqnarray}
We use the image observations, that is, the detected 2D keypoints $\mathbf{x}^i_t$ and confidences $\psi^i_t$~\cite{xu2022vitpose}, to define the joint reprojection loss:
\begin{equation}
\label{eqn:reprojection}
E_{\textrm{data}} = \sum_{i=1}^N \sum_{t=1}^T \ti\psi \rho \big( \Pi_K ( R_t \cdot \w\ti{\tf{J}} + \alpha T_t )- \ti{\tf{x}} \big),
\end{equation}
where $\Pi_K\big(\begin{bmatrix} x_1 & x_2 &x_3 \end{bmatrix}^\top\big) =
K \begin{bmatrix}\frac{x_1}{x_3} & \frac{x_2}{x_3} & 1\end{bmatrix}^\top$
is perspective camera projection
with camera intrinsics matrix $K \in \mathbb{R}^{2 \times 3}$,
and $\rho$ is the robust Geman-McClure function~\cite{geman1987statistical}.

In the first stage of optimization, we align the parameters of the people in the world with the observed 2D keypoints.
Because the reprojection loss (\ref{eqn:reprojection}) is very under-constrained, 
in this stage, we optimize only the global orientation and root translation $\{ \w\ti{\Phi}, \w\ti{\Gamma} \}$
of the human pose parameters:
\begin{equation}
    \min_{\{\{^\textrm{w}\ti{\Phi}, ^\textrm{w}\ti{\Gamma}\}_{t=1}^T \}_{i=1}^N}
    \lambda_{\textrm{data}} E_{\textrm{data}}.
    \label{eq:root-fit}
\end{equation}
We optimize Equation~\ref{eq:root-fit} for 30 iterations with $\lambda_\textrm{data}=0.001$.

\subsection{Smoothing trajectories in the world}
\label{sec:stage2}
We next begin optimizing for the camera scale $\alpha$ and the human shape $\beta_i$ and body pose $\ti\Theta$ parameters.
As we begin to update $\alpha$, we must disambiguate
the contribution of camera motion $\{R_t, \alpha T_t\}$
from the contribution of human translation $\ti{\Gamma}$
to the reprojection error of the joints in Equation~\ref{eqn:reprojection}.
To do this, we introduce additional priors about how humans move in the world to constrain the displacement of the people to be plausible.
We ultimately use an data-driven transition-based human motion prior in our final stage of optimization;
to prepare for this, we perform an optimization stage to smooth the transitions between poses in the world trajectories.
We use a simple prior of joint smoothness, or minimal kinematic motion:
\begin{eqnarray} \label{eqn:smoothness}
E_{\textrm{smooth}}= \sum_i^N \sum_t^T \| \mathbf{J}^i_t - \mathbf{J}^i_{t+1} \|^2.
\end{eqnarray}
We also use priors on shape~\cite{bogo2016keep} $E_{\beta} = \sum_i^N \|\beta^i\|^2$ and pose $E_{\textrm{pose}}=\sum_{i=1}^N \sum_{t=1}^T \|\zeta_t^i\|^2$, where $\zeta_t^i \in \mathbb{R}^{32}$ is a representation of the body pose parameters $\Theta_t^i$ in the latent space of the VPoser model~\cite{pavlakos2019expressive}.
We add these losses to Equation~\ref{eq:root-fit}, and optimize over the entire $\w \ti{\bigP}$ and camera scale $\alpha$:
\begin{equation}
    \min_{\alpha, \{\{^\textrm{w}\ti{\bigP}\}_{t=1}^T \}_{i=1}^N}
    \lambda_\textrm{data} E_\textrm{data}
    + \lambda_\beta E_{\beta} + \lambda_{\textrm{pose}} E_{\textrm{pose}}
    + \lambda_\textrm{smooth} E_{\textrm{smooth}}.
    \label{eq:smooth-fit}
\end{equation}
We optimize for 60 iterations and use $\lambda_\textrm{smooth} = 5, \lambda_\beta=0.05, \lambda_\textrm{pose}=0.04$.
\subsection{Incorporating learned human motion priors}
\label{sec:stage3}
We finally introduce a learned motion prior that better captures the distribution of plausible human motions.
We use the transition-based motion prior, HuMoR~\cite{rempe2021humor},
in which the likelihood of a trajectory $\{\tf{s}_0, \dots, \tf{s}_T\}$
can be factorized into the likelihoods of transitions between consecutive states, $p_\theta(\tf{s}_t | \tf{s}_{t-1})$,
where $\tf{s}_t$ is an augmented state representation used by~\cite{rempe2021humor},
containing the SMPL pose parameters $\tf{P}_t$, as well as additional velocity and joint location predictions.
The likelihood of a transition $p_\theta(\tf{s}_t | \tf{s}_{t-1})$ 
is modeled by a conditional variational autoencoder (cVAE) as
\begin{equation*}
    p_\theta(\tf{s}_t | \tf{s}_{t-1}) = \int_{\tf{z}_t} 
    p_\theta(\tf{z}_t | \tf{s}_{t-1})
    p_\theta(\tf{s}_t | \tf{z}_{t}, \tf{s}_{t-1}),
\end{equation*}
where $\tf{z}_t \in \mathbb{R}^{48}$ is a latent variable representing the transition between $\tf{s}_{t-1}$ and $\tf{s}_t$.
The conditional prior $p_\theta(\tf{z}_t | \tf{s}_{t-1})$
is parameterized as a Gaussian distribution with learned 
mean $\mu_\theta(\tf{s}_{t-1})$ and covariance $\sigma_\theta(\tf{s}_{t-1})$.
We then use this learned prior in an energy term on the latents $\ti{\tf{z}}$:
\begin{equation}
\label{eqn:cvae}
    E_\textrm{CVAE} = -\sum_i^N\sum_t^T
    \log \mathcal{N}(\mathbf{z}_{t}^i; \mu_\theta(\tf{s}^i_{t-1}),\sigma_\theta(\tf{s}^i_{t-1})).
\end{equation}

We perform optimization over the initial states $\tf{s}^i_0$,
the camera scale $\alpha$,
and latent variables $\ti{\tf{z}}$, for timesteps $t=1,\dots, T-1$ and people $i=0,\dots,N-1$.
We initialize the transition latents $\ti{\tf{z}}$ from consecutive states $\tf{s}^i_{t-1}$ and $\ti{\tf{s}}$
with the pre-trained HuMoR encoder $\mu_\phi$,
and use the HuMoR decoder $\Delta_\theta$ to recursively roll out state $\tf{s}^i_t$ from the previous state $\tf{s}^i_{t-1}$ and current latent $\ti{\tf z}$:
\begin{align}
\label{eq:humor_z_init}
    &\ti{\tf{z}} = \mu_\phi (\tf{s}^i_t, \tf{s}^i_{t-1}),
    &\ti{\tf{s}} = \tf{s}^i_{t-1} + \Delta_\theta(\ti{\tf{z}}, \tf{s}^i_{t-1}).
\end{align}
We recover the entire trajectories $(\tf{s}^i_0,\dots,\tf{s}^i_T)$ for all people $i$
by autoregressively rolling out the initial states $\tf{s}^i_0$ with the latents $\ti{\tf{z}}$ initialized in Eq.~\ref{eq:humor_z_init}.
We also carry over additional losses $E_\textrm{stab}$ from~\cite{rempe2021humor} to %
regularize the predicted velocity and joint location components of $\tf{s}^i_t$ to be physically plausible
and consistent with the pose parameter components of $\tf{s}^i_t$; please see \cite{rempe2021humor} for more details.
We denote all prior optimization terms as $E_\textrm{prior} = \lambda_\textrm{CVAE} E_\textrm{CVAE} + \lambda_\textrm{stab} E_\textrm{stab}.$

Following~\cite{rempe2021humor}, we also optimize for the ground plane of the scene $g \in \mathbb{R}^3$,
and use the decoder to predict the probability of ground contact $c(j) \in [0, 1]$ for joints $j$.
We enforce a zero velocity prior on joints that are likely to be in contact with the ground $g$
to prevent unrealistic foot-skate:
\begin{equation}
E_{\textrm{skate}} = \sum_i^N \sum_t^T \sum_j^J c^i_t(j) \|\mathbf{J}^i_t(j) - \mathbf{J}^i_{t+1}(j)\|,
\end{equation}
while also encouraging their distance from the ground to be less than a threshold $\delta$:
\begin{equation}
E_{\textrm{con}} = \sum_i^N \sum_t^T \sum_j^J c^i_t(j) \max (d(\mathbf{J}^i_t(j), g)-\delta,0).
\end{equation}
Here, $d(\mathbf{p}, g)$ defines the distance between point $\mathbf{p} \in \mathbb{R}^3$ and the plane $g \in \mathbb{R}^3$,
and we optimize $g$ as a free variable shared across all people and timesteps.
We denote these constraints as $E_{\textrm{env}} = \lambda_\textrm{skate} E_{\textrm{skate}} + \lambda_\textrm{con} E_{\textrm{con}}$. 

Our optimization problem for this stage is then
\begin{equation} \label{eq:final_objective}
\begin{split}
    \min_{ \substack{\alpha , g, \{\tf{s}^i_0\}_{i=1}^N, \\ \{ \{ \ti{\tf{z}} \}_{t=1}^T \}_{i=1}^N} }
    \lambda_\textrm{data} &E_\textrm{data}
    + \lambda_\beta E_{\beta} + \lambda_{\textrm{pose}} E_{\textrm{pose}} \\[-4ex]
    &+ E_\textrm{prior} + E_\textrm{env}.
\end{split}
\end{equation}
We optimize Equation~\ref{eq:final_objective} with an incrementally increasing horizon,
increasing $T$ in chunks of 10: $H = 10\tau$, $\tau = 1, \dots, \lceil \frac{T}{10} \rceil$.
We optimize $\{\tf{z}_0,\dots,\tf{z}_H\}$ adaptively, 
rolling out the trajectory by 10 more frames each time the loss decreases less than a threshold $\gamma$,
for a minimum of 5 iterations and maximum 20 iterations.
We use $\lambda_\textrm{CVAE} = 0.075, \lambda_\textrm{skate}=100,$ and $\lambda_\textrm{con}=10$.
We perform all optimization with PyTorch~\cite{paszke2019pytorch} using the L-BFGS algorithm with learning rate 1.

\input{sec/3d_method}

%% file: sec/3d_method.tex
\subsection{Implementation details}
\noindent \textbf{Missing observations:}
Our approach fills in missing information for every person caused by poor detection and/or occlusion. 
To initialize the person variables for these frames, \ie, $\Phi_t^i, \Theta_t^i, \Gamma_t^i$, we interpolate $\Phi_t^i, \Theta_t^i$ and on $SO(3)$ and $\Gamma_t^i$ in $\mathbb{R}^3$. While no data term is available for these missing frames, motion priors along with the camera motion provide additional signals to these parameters. 

\noindent \textbf{Handling multiple people in-the-wild:}
Although simple in concept, reasoning about multiple people at once in the already big optimization problem is a challenge, particularly since in videos in-the-wild, not all people appear at the same timestamp.
People can enter the video at any frame, leave and come back again.
Our implementation is designed to handle these cases well.
We also use an improved version of PHALP with a stronger detector~\cite{li2022exploring}, which we refer to as PHALP+.
Please see the appendix for more details.
Code is available at the \href{https://vye16.github.io/slahmr}{project page}.

%% file: sec/4_experiments.tex
We demonstrate quantitatively and qualitatively that our approach effectively reconstructs human trajectories in the world.
We also demonstrate quantitatively that the camera scale we recover can be used to improve people tracking in videos.
We encourage viewing additional video results on the \href{https://vye16.github.io/slahmr}{project page}.

{\bf Datasets.} 
Datasets typically used for evaluation in the 3D human pose literature generally only provide videos captured with a static camera (\eg, Human3.6M~\cite{ionescu2013human3}, MPI-INF-3DHP~\cite{mehta2017monocular}, MuPoTS-3D~\cite{mehta2020xnect}, PROX~\cite{hassan2019resolving}).
3DPW~\cite{von2018recovering} is a dataset captured with moving cameras, and includes indoor and outdoor sequences of people in natural environments.
However, as is also discussed by previous work~\cite{yuan2022glamr,zhang2021learning},
it only provides 3D pose ground truth in the local frame of the person,
and it is not possible to evaluate the global motion of the person.
We use only 3DPW to perform ablations on the reconstructed local pose using our method.

The most relevant dataset that is captured with dynamic cameras and provides ground truth 3D pose in the global frame is the recently introduced {\bf EgoBody dataset}~\cite{zhang2021egobody}.
EgoBody is captured with a head-mounted camera on an {\it interactor},
who sees and interacts with a second {\it interactee}.
The camera moves as the interactor moves,
and the ground truth 3D poses of the interactee are recorded in the world frame.
Because videos are recorded from a head-mounted camera,
Egobody videos have heavy body truncations, with the interactee often only visible from chest or waist up.

\begin{figure*}
\centering
\includegraphics[width=\linewidth]{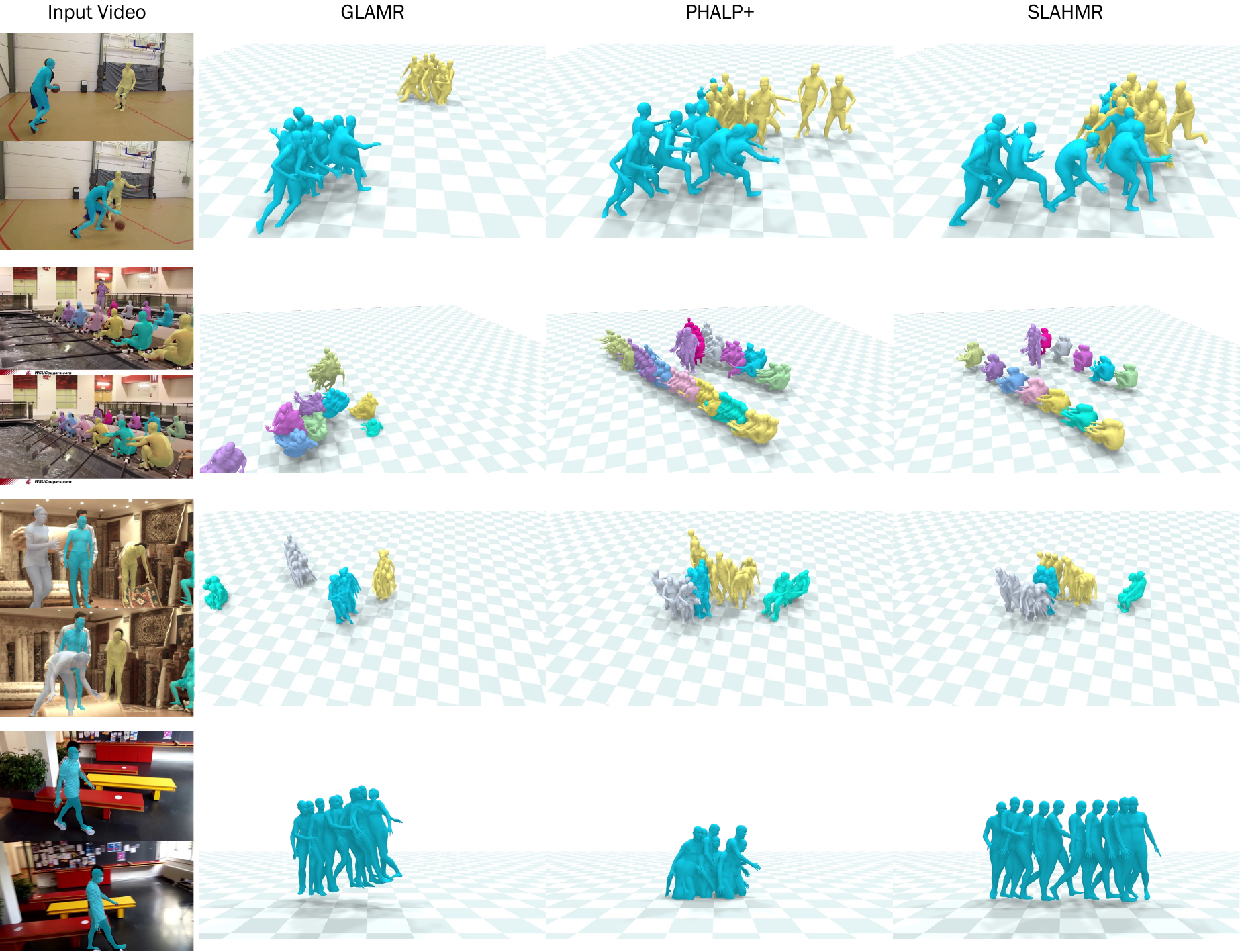}
\caption{
\textbf{Qualitative results of the proposed approach.} The first three rows are from PoseTrack~\cite{andriluka2018posetrack} and the last row is from EgoBody~\cite{zhang2021egobody}. The columns compare results on three approaches: GLAMR~\cite{yuan2022glamr}, PHALP$^+$~\cite{rajasegaran2022tracking} which is the input to our system, and SLAHMR. We visualize the top-down view of the recovered motion trajectory across the entire frames. Note that GLAMR struggles to recover a plausible global trajectory (first and second row)  and also struggles on a sequence with close to static camera input (third row). Our optimization improves upon the inputs in reducing the jitter and recovering plausible motion trajectory with more plausible depth relationship between the people. Please the \href{https://vye16.github.io/slahmr}{project page} for the results in video format. 
}
\label{fig:big}
\end{figure*}

We also demonstrate our approach on the PoseTrack dataset~\cite{andriluka2018posetrack}.
PoseTrack is an extremely challenging in-the-wild dataset originally designed for tracking.
It spans a wide variety of activities, involving many people with heavy occlusion and  interaction.  %
We use PoseTrack to qualitatively demonstrate the robustness of our method,
and show many results in Figure~\ref{fig:big} and in the Sup. Video.
Because there is no 3D ground truth on PoseTrack, we perform quantitative evaluation through the downstream task of tracking.  
We show that
reasoning about the tracks with the scaled camera trajectory
recovered by our approach, can boost its state-of-the-art performance on PoseTrack. %

\input{tab/ablation_ego}
\input{tab/ablation_3dpw}

{\bf Evaluation metrics:} %
We report a variety of metrics, with a focus on metrics that compute the error on the {\it world} coordinate frame.
World PA Trajectory - MPJPE (WA-MPJPE) reports MPJPE~\cite{ionescu2013human3} after aligning the \textit{entire trajectories} of the prediction and ground truth with Procrustes Alignment.
World PA First - MPJPE (W-MPJPE) reports the MPJPE after aligning the \textit{first frames} of the prediction and the ground truth.
PA-MPJPE reports the MPJPE error after aligning \textit{every frame} of the prediction and the ground truth.
We also report Acceleration Error computed as the difference between the magnitude of the acceleration vector at each joint.
Please see the Sup. Mat.for more details on the evaluation protocol.
For tracking, we report identity switch metrics; 
other commonly used tracking metrics measure quality of association and detection,
but we use the same detection and association protocol in all baselines, so we omit those.

\subsection{Egobody results}
\input{tab/baselines_ego}
To demonstrate the effect of the different components of our system, we first perform an ablation study reporting results in Table~\ref{tbl:egobody_ablation}.
We use the metrics presented earlier and we discuss different settings.
We start with the result of the full system and remove some key components.
We report performance metrics of
(i) our method without the last stage of optimization, \ie, without the motion prior and scale (``w/o last stage"),
(ii) our method before optimization in the world, \ie, only the PHALP$^+$ results with the estimated cameras, and
(iii) the basic results of PHALP$^+$ in the camera frame, without estimated cameras at all.
We report metrics on the reconstructed trajectories in the world (W-MPJPE, WA-MPJPE, Acc Error) in Table~\ref{tbl:egobody_ablation}.

For completeness, we also report the PA-MPJPE, which is common in the literature,
for the same ablations on both the Egobody and 3DPW datasets.
Because 3DPW annotates up to two people's poses for the captured sequences, we evaluate two variants of our method.
3DPW${}^\dagger$ uses our full system's pre-processing: PHALP$^+$ to detect, track, and estimate people's initial 3D local poses.
3DPW${}^*$ uses each person's ground-truth tracks, and runs PHALP$^+$'s 3D pose estimation only.
We note that 3DPW${}^*$, \ie, using the ground truth tracks, is most similar to the current evaluation practices on 3DPW.
We report the results in Table~\ref{tbl:3dpw_ablation}.
We see that in Egobody, in which the subject is often heavily truncated, the prediction method PHALP$^+$ achieves better performance.
In other words, further optimization with truncated observations can reduce performance;
this is a known issue for mesh recovery methods~\cite{joo2021exemplar,kocabas2021pare,pavlakos2022human}.
However, in 3DPW, in which the subjects are more fully visible, our method improves upon the initial predictions from PHALP$^+$.
Ultimately, PA-MPJPE only captures local pose accuracy, and cannot describe the global attributes of the full trajectories.

We also compare our approach with a series of state-of-the-art methods for human mesh recovery in Table~\ref{tbl:egobody_sota}.
The closest to our system is GLAMR~\cite{yuan2022glamr}, which also estimates 3D body reconstructions in the world frame.
As we see in Table~\ref{tbl:egobody_sota}, we comfortably outperform GLAMR.
Because GLAMR computes the world trajectory based on local pose estimates alone,
it is especially sensitive to the extreme truncation in Egobodoy videos.
In contrast to GLAMR, we leverage the relative camera motion to achieve significantly better reconstruction results, globally and locally.

We also compare against human mesh recovery baselines that compute the motion in the camera frame only.
We include state-of-the-art baselines for
a) single frame mesh regression (PHALP~\cite{rajasegaran2022tracking}),
b) temporal mesh regression (VIBE~\cite{kocabas2020vibe}),
and c) temporal mesh optimization (VIBE-opt~\cite{kocabas2020vibe}).
Our method outperforms all baselines in world-level metrics, and is second best in local pose metrics.

\begin{figure*}[!ht]
    \centering
    \includegraphics[width=0.98\textwidth]{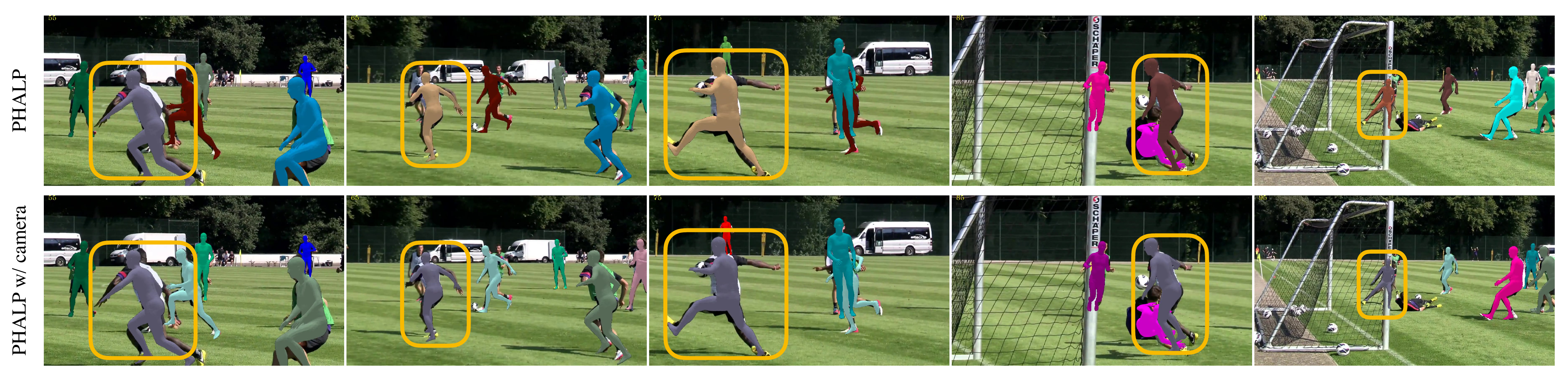}
    \vspace{-0.2cm}
    \caption{\textbf{Effect of the estimated camera for tracking.} We integrate our predicted camera with the PHALP tracking system~\cite{rajasegaran2022tracking} and we show the effect on the PoseTrack dataset~\cite{andriluka2018posetrack}.
    Explicit modeling of the camera makes PHALP tracking robust to abrupt camera motions.
    Notice how out-of-the-box PHALP (top row) leads to multiple identity switches, while when we integrate the camera (bottom row), we can achieve consistent tracking over the duration of the video (\ie, the different people maintain the color indicating their identity).}
    \label{fig:single_ve_multi_pose}
    \vspace{-0.4cm}
\end{figure*}

\subsection{Posetrack results}
For the PoseTrack dataset, we show qualitative results in Figure~\ref{fig:big},
comparing against GLAMR and PHALP$^+$ (an input to our method).
We also provide qualitative reconstructions of all Posetrack validation videos in the supplemental video at the \href{https://vye16.github.io/slahmr}{project page}.
We highly encourage seeing the qualitative results in video to appreciate the improvements in world trajectories.
We see that our method recovers smoother trajectories that are more consistent with the dynamic scene in the input videos.
We see that GLAMR struggles to plausibly position multiple people in the same world frame.

We also demonstrate a downstream application of our method, by providing helpful camera information to the PHALP tracking method~\cite{rajasegaran2022tracking}.
In brief, PHALP uses an estimate of the 3D person location in the \textit{camera} frame for tracking.
We posit that tracking is better done in the world coordinate frame, as it will be invariant to camera motion. 
We provide to PHALP the recovered camera motion from our approach, \ie, relative cameras from~\cite{teed2021droid} with the scale factor $\alpha$ from our optimization to place the people in the world coordinate frame. We make minimal adaptation to the PHALP algorithm to demonstrate the effect of camera information, however there is a potential for even more improvement.  Please see the supplemental for more details. The rest of the tracking procedure operates as in PHALP. %
We report the results in Table~\ref{tbl:posetrack_sota}.
We report both PHALP and PHALP$^+$, which uses a better detection system~\cite{li2022exploring},
along with two variants using additional camera information:
(i) using the cameras of~\cite{teed2021droid}, without rescaling to the scale of human motion,
and (ii) also using the recovered scale $\alpha$ from our optimization.
We see that using out-of-the-box cameras from~\cite{teed2021droid} does not change performance.
In contrast, using the recovered scale from our approach makes a significant improvement to the ID switch metric.
This observation indicates our method recovers a more accurate scale, and shows the benefit it can have in the challenging tracking scenario.
In Figure~\ref{fig:single_ve_multi_pose}, we also demonstrate an example of the better behavior we achieve with our tracking.

\begin{table}[!h]
\begin{center}
\begin{tabular}{l l}
\toprule[0.4mm]
Method & IDs$\downarrow$ \\ 
\midrule
T3DP~\cite{rajasegaran2021tracking} & 655 \\
PHALP~\cite{rajasegaran2022tracking} & 541 \\
\midrule
PHALP$^+$ & 450 \\
PHALP$^+$ + \cite{teed2021droid} cams & 446$\ \textcolor{teal}{(-0.8\%)}$ \\
PHALP$^+$ + \cite{teed2021droid} cams + $\alpha$ & 420$\ \textcolor{teal}{(-6.7\%)}$ \\
\bottomrule[0.4mm]
\end{tabular}
\end{center}
\vspace{-0.4cm}
\caption{{\bf Effect of tracking with camera information on PoseTrack~\cite{andriluka2018posetrack}.}
We start with PHALP$^+$,
which is PHALP~\cite{rajasegaran2022tracking} using a more accurate detector~\cite{li2022exploring}.
Directly using cameras from~\cite{teed2021droid} yield a small difference in ID switches.
However, using the scale $\alpha$ we recover, yields a significant improvement,
and highlights the benefit of integrating reliable camera information for tracking.
}
\label{tbl:posetrack_sota}
\vspace{-0.5cm}
\end{table}

%% file: tab/ablation_ego.tex
\begin{table}[t]
\begin{center}
\resizebox{\linewidth}{!}{
\begin{tabular}{lrrr}
\toprule
Method  &
W-MPJPE$\downarrow$ &
WA-MPJPE$\downarrow$ &
Acc Error$\downarrow$ \\
\midrule
Full system
& \textbf{141.1} & \textbf{101.2} & \textbf{25.78} \\
w/o last stage
& 234.0          & 152.9          & 275.9         \\
PHALP$^+$ w/ SfM
& 253.6          & 150.3          & 302.1         \\
PHALP$^+$ ~\cite{rajasegaran2022tracking}
& 387.8          & 204.9          & 307.6        \\      
\bottomrule
\end{tabular}
}
\end{center}
\vspace{-0.2cm}
\caption{
{\bf Ablation of the proposed system on EgoBody~\cite{zhang2021egobody}.}
Removing any of the proposed components has a significant effect on the final performance of the system.
We report W-MPJPE and WA-MPJPE in mm, and Acc Error in mm/s$^2$.
Note the significant difference motion prior and scale makes in the acceleration error.
}
\label{tbl:egobody_ablation}
\end{table}

%% file: tab/ablation_3dpw.tex
\begin{table}[t]
\begin{center}
\begin{tabular}{lrrrr}
\toprule
Method  &
Egobody &
3DPW$^\dagger$ &
3DPW$^*$ \\
\midrule
Full system
& 79.13          & \textbf{62.60}   & \textbf{55.86} \\
w/o last stage
& 88.18          & 64.50            & 59.19         \\
PHALP$^+$ ~\cite{rajasegaran2022tracking}
& \textbf{72.16}  & 64.68            & 56.70         \\
\bottomrule
\end{tabular}
\end{center}
\vspace{-0.2cm}
\caption{{\bf Comparison of PA-MPJPE for Egobody and 3DPW.}
All errors in mm. 3DPW$^\dagger$ uses the detected PHALP tracks that best match the ground truth track, the result if the system is run out of the box.
3DPW$^*$ uses the ground truth person tracks and is most comparable to existing evaluation protocols.
}
\label{tbl:3dpw_ablation}
\end{table}

%% file: tab/baselines_ego.tex
\begin{table}[t]
\begin{center}
\resizebox{\linewidth}{!}{
\begin{tabular}{lrrrr}
\toprule
Method &
W-MPJPE$\downarrow$ &
WA-MPJPE$\downarrow$ &
Acc Err$\downarrow$ &
PA-MPJPE$\downarrow$  
\\
\midrule
PHALP$^+$~\cite{rajasegaran2022tracking}
& 387.8          & 204.9          & 307.6         & \textbf{72.16}  \\
VIBE~\cite{kocabas2020vibe}
& 500.4          & 259.5          & 524.2         & 100.5   \\
VIBE-opt~\cite{kocabas2020vibe}
& 453.2          & 246.0          & 481.1         & 100.4   \\
\midrule
GLAMR~\cite{yuan2022glamr}
& 416.1          & 239.0          & 173.5         & 114.3   \\
\textbf{SLAHMR}
& \textbf{141.1} & \textbf{101.2} & \textbf{25.78} & 79.13    \\
\bottomrule
\end{tabular}
}
\end{center}
\vspace{-0.4cm}
\caption{{\bf Comparison with the state of the art on EgoBody dataset~\cite{zhang2021egobody}.}
We compare our approach with a variety of state-of-the-art methods for human mesh recovery.
GLAMR is the only other approach that attempts to recover the global human motion trajectory,
but does so from only local pose transitions, without considering scene pixel motion.
Our approach obtains significant improvement in the world trajectory metrics, as well as in the acceleration error.}
\label{tbl:egobody_sota}
\end{table}

%% file: sec/6_conclusion.tex
\looseness=-1 
We propose a method for recovering human motion trajectories in the world coordinate frame from challenging videos with moving cameras. Our approach leverages relative camera estimates from scene pixel motion to optimize trajectories jointly with learned human motion priors for all people in the scene. This allows us to outperform state-of-the-art methods on the Egobody dataset and generate plausible trajectories for scenes with multiple people and challenging camera motions, as demonstrated by our experiments on the PoseTrack dataset.

While our system unlocks many new sources of human data, many problems remain to be addressed.
In-the-wild videos often have ill-posed multiview geometry,
such as predominantly rotational camera motion or co-linear motion between humans and cameras.
Our method can recover inconsistent trajectories in these cases.
Please see the supplemental video for examples.
An exciting avenue for future work would be to incorporate human motion priors to also constrain and update the camera and scene reconstruction.

\noindent
{\bf Acknowledgements:} This research was supported by the DARPA Machine Common Sense program as well as BAIR/BDD sponsors, the Hellman Fellows Partnership, 

%% file: sec/7_supp.tex
\section{Details of EgoBody evaluation}
In Section 4.1 of the main manuscript, we present an experiment on the EgoBody dataset. Here, we provide more details about this evaluation.

We report results on the validation set of EgoBody.
Regarding the estimated camera, we use DROID-SLAM~\cite{teed2021droid} with ground truth intrinsics.
Regarding the person of interest, we first use PHALP$^+$~\cite{rajasegaran2022tracking} (which is the same with out-of-the-box PHALP, but with a more robust detection system~\cite{li2022exploring}), on each sequence.
Since there may be multiple people in the frame (but the dataset provides 3D ground truth only for one main person), we then associate the inferred tracklets with %
the person of interest with the 3D ground-truth pose.
For each detected bounding box, we run a 2D keypoint detection network~\cite{xu2022vitpose}.
We run our method and our baselines~\cite{kocabas2020vibe, yuan2022glamr} on the detected tracklets using the same detections (bounding box, 2D keypoints) and ground-truth intrinsics.
To accelerate inference, we split the original videos on sequences of 100 frames and we optimize each sequence separately.
We report results using both local pose metrics, \ie, PA-MPJPE~\cite{kanazawa2018end} and global metrics that consider the global estimated trajectory across the whole reconstructed sequence. More specifically, we report results in two settings a) after aligning the predicted sequence with the ground truth sequence using Procrustes (World PA Trajectory - MPJPE), and b) after aligning the first frame of the predicted sequence with the first frame of the ground truth sequence using Procrustes (World PA First - MPJPE).

\section{Details of PoseTrack tracking experiment}
In Section 4.2 of the main manuscript, we present an ablation where we leverage the estimated camera and the optimized scale $\alpha$ for the purposes of tracking on the PoseTrack dataset~\cite{andriluka2018posetrack}. Here, we give more details about this implementation.

To make a direct comparison with PHALP~\cite{rajasegaran2022tracking}, we make minimal modifications to the main algorithm.
PHALP uses four cues; appearance, pose, 2D location and nearness of the person.
We did not modify the appearance and the pose cues, but only applied the effect of the camera on the location cues, \ie, 2D location and nearness.
More specifically, PHALP estimates the 3D location for each person detection in the camera frame, using a single-frame HMR model~\cite{kanazawa2018end}.
Given our estimated camera for each frame (\ie, relative camera from~\cite{teed2021droid} and estimated world scale $\alpha$ from our optimization), we first transform PHALP's 3D location to the world frame (\ie, coordinate frame of the first video frame).
Next, PHALP projects these 3D location to the image plane, keeping track of the 2D location, while also recording the depth (nearness) as a separate feature.
For simplicity, we take the $(X,Y,Z)$ location of each detection in the world frame and, a) keep the $(X,Y)$ part of the location of each detection to represent the 2D location, (after normalizing it to $[0, 1]$, the same way that PHALP does) and b) use the $Z$ coordinate to compute the nearness.
The rest of the pipeline remains the same as PHALP. Essentially, the only difference is that the location of the people are considered in the world coordinate instead of the camera coordinate frame. 

We highlight that we only make minimal adaptations to the main PHALP algorithm to demonstrate the effect of camera information for tracking, but there is further room for improvement.
For example, considering that we have access to the explicit 3D location for each detection in the world frame, we could also explore tracking using 3D location as a cue, instead of splitting the position cue to 2D location and depth/nearness, but this would require modification to the PHALP's tracking parameters. %
Similarly, we could leverage our optimized results to compute more reliable affinity metrics on the pose, but here our goal was to decouple the benefit of the better camera from other cues, \ie, our more stable pose.
It would be an interesting direction for future work to integrate all these updates and implement a more robust tracking system using information for camera motion.

\section{Additional implementation details}

\paragraph{Floor specification:}
When multiple people are on the same floor level, our optimization becomes better constrained because all of them need to share the same floor $g$, meaning that the motion of more people provides constraints for the optimization of the $g$ variable.
However, in many real world videos, people are in different floor levels.
In that case, when we observe that it is not possible to solve Equation~\ref{eq:final_objective} with a single floor variable $g$, we separate the people in $K$ clusters based on the locations of their feet, and introduce $K$ separate floor variables $g^k$.
The people in cluster $k$ shares the same floor $g^k$ and the optimization continues as usual.

\paragraph{Handling multiple people}
A distinct challenge of in-the-wild videos is properly handling multiple person tracks of undetermined length as they undergo occlusion.
During the first two stages of optimization, each person's pose is optimized independently.
During these stages, we only optimize the people that are visible, and mask out losses on the predictions of any frame and any track that are not visible.

During the last stage,
optimizing all tracks in a single batch allows scale and ground contact information to be shared between people.
To do this in our incremental optimization scheme (described in Section 3.4 of the main text),
we store each track with respect to its \textit{first appearance}, rather than with respect to the first frame of the video.
We pad the end of each track to be $T_{\text{max}}$, the length of the longest track.
Specifically, for each track, we store the start and end times of the track, $(t_{\text{start}}, t_{\text{end}})$, and latent vectors $z_{0:T_\text{max}}$.
The latents of each track are contiguous in time (we infill occlusions between the first and last appearances), but do not all start or end at the same timestep.

In an optimization step at the rollout horizon $\tau$, we roll out $10\tau$ steps of each track $X_{0:10\tau}$,
where $X$ is the decoded latent state.
We then scatter each track $X_{0:10\tau}$ into the interval $[t_{\text{start}}, \text{min}(t_{\text{end}}, t_{\text{start}} + 10\tau)]$ of input video's timeline.
That is, each state $X_k$ synchronized to the original time $t$ it occurred in, and remove the padded states.
We then only optimize the track over the time segment containing $X_{0:10\tau}$, $[t_{\text{start}}, \text{min}(t_{\text{end}}, t_{\text{start}} + 10\tau))]$, and mask out the frames of each track that fall outside of this interval.

The runtime of optimization grows linearly with the number of people we track.
Optimizing a sequence of around 100 frames and 4 people requires around 40 minutes.

\section{Robustness}
One of our observations with regards to using the HuMoR motion prior~\cite{rempe2021humor} is that it can be challenging to optimize, especially over a long sequence.  %
This results in our decision to optimize the pose sequences of every person in a rollout horizon, as described in the previous section. %
This increases the robustness of the optimization for longer sequences and it should be applicable to any motion prior that also models the transition, \eg,~\cite{ling2020character}.%

Moreover, HuMoR assumes static camera.
When used on sequences with camera movement, without modeling the camera motion as we do, it can lead to catastrophic failures in the optimization.
For example, in Egobody, we observed that HuMoR fails on 30\% of the sequences when we use identity (static) camera.
In contrast to that, our approach, even with imperfect camera motion, \ie, using the estimates from~\cite{teed2021droid} as we do, leads to successful optimization in 99\% of the sequences; for the rare cases where optimization of the HuMoR motion prior fails, we simply revert back to the results of the previous step where we optimize with the smoothness motion prior.

On the more challenging PoseTrack sequences, we also observe some rare optimization failures.
Most of those are related to the single floor assumption and can be addressed by clustering the people in different floors, as described in Section 3.4 of the main manuscript.

\section{Limitations}
One of the limitation of our approach is that we rely on outputs from other methods (\eg, estimated camera from~\cite{teed2021droid} with approximate intrinsics for in-the-wild videos, person tracking from~\cite{rajasegaran2022tracking}), which sometime can propagate failures to our optimization.

For example, SfM approaches often have trouble distinguishing between translational and rotational motions, particularly with large focal length.
Although our optimization can typically infer reasonable motions even with these imperfect camera estimation, an exciting future work is to jointly optimize the camera motion and human motion, which requires also updating the 3D structure.

Another failure mode is in case of identity switch errors in tracking, with the most harmful being errors that merge two different people into a single tracklet.
Although we do not explicitly reason about tracklet identity during our optimization, we provide an experiment where PHALP makes better use of information about camera motion (main manuscript, Section 4.2).
Future work could also solve the association problem while optimizing over people and camera's motion.

Finally, we observed some inherently challenging motions to decouple from a monocular video, \eg, when people move co-linearly with the camera. 
In these cases, our approach can underestimate the location evolution of the people, \eg, causing people to run in the same location. Please see the example in the supplemental video. 
In these situations, future work could consider also priors for the background scale, \eg, by using monocular depth cues~\cite{ranftl2020towards}, which could help to better constrain the scale factor $\alpha$.